\documentclass{article}

\usepackage[preprint, nonatbib]{neurips_2022}

\usepackage[utf8]{inputenc} 
\usepackage[T1]{fontenc}    
\usepackage{hyperref}       
\usepackage{url}            
\usepackage{booktabs}       
\usepackage{amsfonts}       
\usepackage{nicefrac}       
\usepackage{microtype}      
\usepackage{xcolor}         
\usepackage{graphicx}       
\usepackage{amsmath}
\usepackage{multirow}

\title{Language Does More Than Describe: On The Lack Of Figurative Speech in Text-To-Image Models}

\author{%
  Ricardo Kleinlein\\
   \And
   Cristina Luna-Jiménez \\
   \AND
   Fernando Fernández-Martínez \\
   Grupo de Tecnología del Habla y Aprendizaje Automático\\
  Information Processing and Telecommunications Center\\ 
  E.T.S.I. de Telecomunicación, 
  Universidad Politécnica de Madrid
  28040 Madrid, Spain \\
   \texttt{\{ricardo.kleinlein, cristina.lunaj, fernando.fernandezm\}@upm.es} \\
}

\begin{document}

\maketitle

\begin{abstract}

The impressive capacity shown by recent text-to-image diffusion models to generate high-quality pictures from textual input prompts has leveraged the debate about the very definition of art.
Nonetheless, these models have been trained using text data collected from content-based labelling protocols that focus on describing the items and actions in an image, but neglect any subjective appraisal.
Consequently, these automatic systems need rigorous descriptions of the elements and the pictorial style of the image to be generated, otherwise failing to deliver.
As potential indicators of the actual artistic capabilities of current generative models we characterise the sentimentality, objectiveness and degree of abstraction of publicly available text data used to train current text-to-image diffusion models.
Considering the sharp difference observed between their language style and that typically employed in artistic contexts, we suggest generative models should incorporate additional sources of subjective information in their training in order to overcome (or at least to alleviate) some of their current limitations, thus effectively unleashing a truly artistic and creative generation.

\end{abstract}

\section{The importance of figurative speech in art}

Deep Generative Models (DGM) have become a particularly hot research topic within the field of machine learning~\cite{Oussidi2018}.
DGMs are neural networks with thousands of parameters trained to approximate probability distributions from the observation of a large number of samples, after which they can be used to create new instances that resemble the training data.
Their popularity has been steadily increasing in the last years, reaching the public sphere due to the success of
autoregressive models~\cite{Salimans2017PixeCNN, Oord2016}, Variational AutoEncoders (VAEs)~\cite{Kingma2013, Kingma2019} and Generative Adversarial Networks (GANs)~\cite{Goodfellow2016}.
However, to condition the generation process towards more elaborate image semantics has been a long-standing problem~\cite{Gu2019, Casanova2021, Mirza2014, Reed2016, Park2019}.
With the advent of Large Language Models (LLMs)~\cite{Raffel2020, Brown2020} and CLIP~\cite{Radford2021}, diffusion models can now be conditioned by textual prompts in natural language~\cite{Ho2020}.
Still, some of the most popular ones such as DALL-E2~\cite{Ramesh2022}, Stable Diffusion~\cite{Rombach2021} and Imagen~\cite{Saharia2022} need these prompts to describe as exactly as possible both the items and the pictorial style of the image to generate.

It is not a minor issue, given how tightly bounded human communication and art are.
Even in indigenous cultures~\cite{Brown2018}, art is used to hand information about the feelings, ideas and perception of the artist to an audience, arousing emotions along the process~\cite{tolstoi, Mastandrea2019}.
People often describe artworks in terms associated with emotional states~\cite{carroll2010art}, and as some theorists of art have noted, the exact content of a visual artwork is neither arbitrary nor unimportant, but perfectly aligned with the abstract theme the artist seeks to communicate~\cite{arnheim1954art}.
Furthermore, the elements in an artwork do not even need to be realistic to achieve its goal~\cite{bell1987art}.
We argue current text-to-image generative models are ignoring this fundamental aspect of the human creative process, hence failing to deliver when the prompts conditioning the generation process of the model is not strictly descriptive (Appendix \ref{append: examples}).

\section{Experimental setup and results}

\begin{figure}
    \centering
    \includegraphics[width=\textwidth]{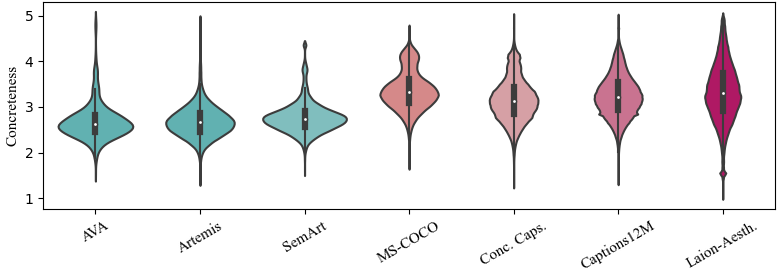}
    \caption{Distributions of average sentence concreteness score in each dataset. The color flips represent the shape of the distribution whereas the black box inside denotes the mean value and standard deviation of the distribution.}
    \label{fig:concreteness-violin}
\end{figure}

Our main argument is that text-to-image generative models are being trained on databases originally created to different purposes like image segmentation, hence limiting the way they interpret human language in the specific context of art generation.
SemArt~\cite{Garcia2019} and Artemis~\cite{achlioptas2021artemis} are examples of datasets specifically aimed at understanding the role played by language in the description of artworks, thus paramount instances of the idiosyncrasy of the language used in artistic contexts.
Similarly, incorporating to our study the descriptions of the challenges contained in the AVA dataset~\cite{Murray2012}, we gain insight into how people ask for new pieces of art (in this case, photography contests). 
Opposed to them, MS-COCO~\cite{Lin2014}, Conceptual Captions~\cite{sharma2018conceptual}, Captions12M~\cite{Changpinyo2021} and Laion-Aesthetics~\cite{Schuhmann2022} are publicly available databases that constitute part of the training material used in text-to-image diffusion models whose language is essentially content-based and aimed to a neutral description of the scene.
We qualitatively describe the sentiment valence~\cite{Hutto2014}, subjectivity~\cite{blob} and average term concreteness~\cite{Brysbaert2014} (the inverse of term abstraction) of image captions following the procedure introduced in \cite{achlioptas2021artemis}.

The most obvious difference can be seen in the average term concreteness per sentence, shown in Figure~\ref{fig:concreteness-violin}.
Apparently, datasets focused on art display more abstract vocabulary.
When we consider their average sentiment valence, they also seem to be shifted towards positive emotions ($\mu=0.27, \sigma^2=0.45$) while content-based data is fundamentally neutral ($\mu=0.1, \sigma^2=0.27$).
Likewise, Artemis, SemArt and AVA present on average a greater ratio of subjective terms ($\mu=0.41, \sigma^2=0.27$) than the rest of datasets ($\mu=0.22, \sigma^2=0.28$).
We also compute the Earth-Moving Distance between each pair of datasets~\cite{Rubner2000}, finding that there are in fact two types of language styles, namely that related to art environments and that of content-based, neutral and objective descriptions (we refer the reader to Appendix \ref{append: eom} for further detail). 

\begin{ack}
The work leading to these results was supported by the Spanish Ministry of Science and Innovation through the projects GOMINOLA (PID2020-118112RB-C21 and PID2020-118112RB-C22, funded by MCIN/AEI/10.13039/501100011033), and AMIC-PoC (PDC2021-120846-C42, funded by MCIN/AEI/10.13039/501100011033 and by the European Union ``NextGenerationEU/PRTR'').
Ricardo Kleinlein’s research was supported by the Spanish Ministry of Education (FPI grant PRE2018-083225).

\end{ack}

\section{Ethical considerations}

The automatic generation of content, be it text, audio or image is now in the middle of a heated debate that may remind us of that happening decades ago when the massive adoption of photography was followed by the discussion about whether it could be considered an artistic expression~\cite{walter}.
Contrarily to cameras, the black-box nature of DGMs makes particularly challenging - although no less important - to understand the impact of the training material of the model on the final outcome to ensure fairness and prevent negative biases.
We hope our work contributes to the demystification of these generative models while planting the seed for future work on their interpretability. 

\bibliographystyle{plain}
\bibliography{references}



\appendix

\section{Instances of prompts from AVA}
\label{append: examples}

\begin{figure}
    \centering
    \includegraphics[width=\textwidth]{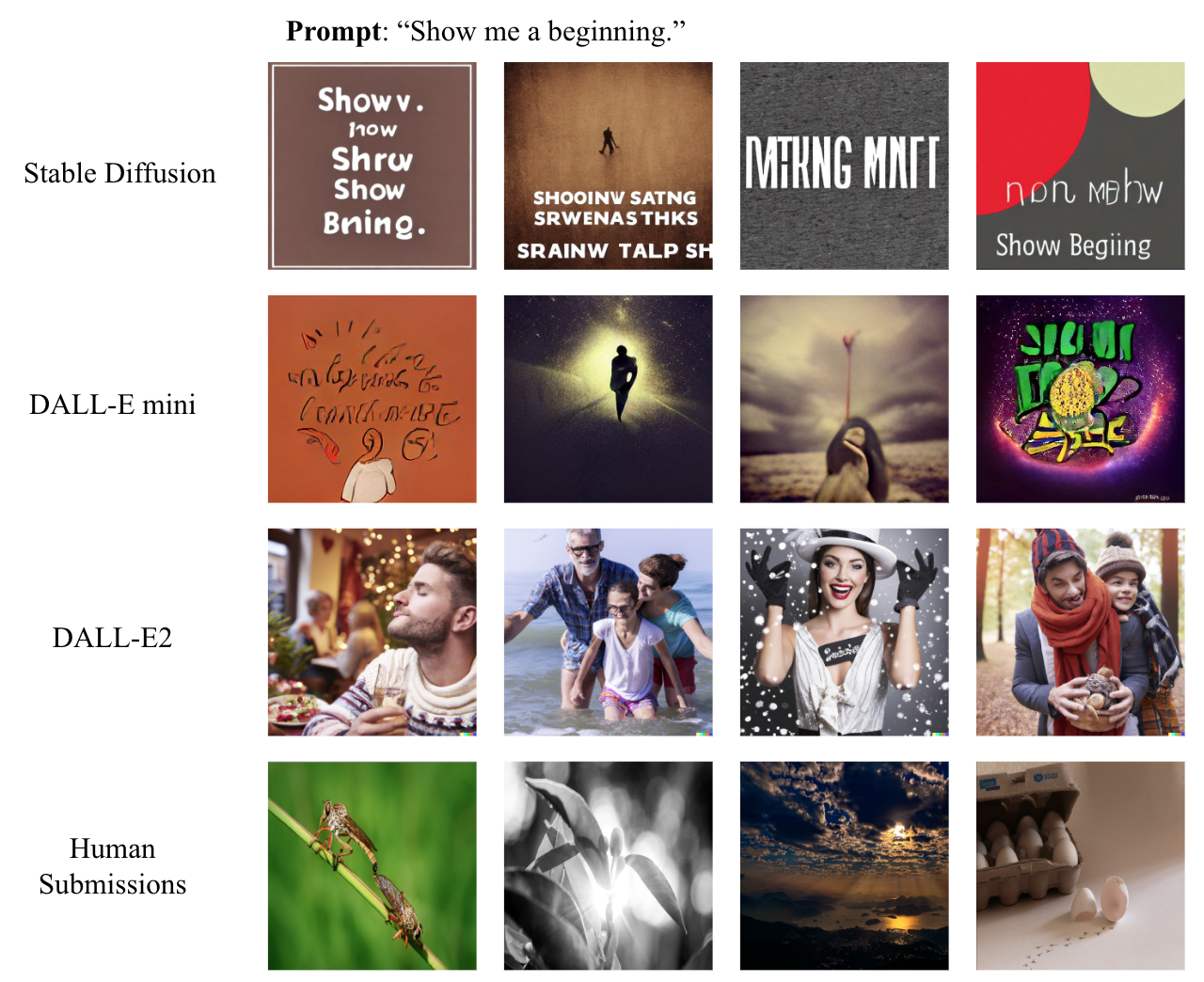}
    \caption{The prompt is deliberately open to interpretation, but these systems do not seem to have captured the underlying abstraction of the term "beginning".}
    \label{fig:example_beginning}
\end{figure}

\begin{figure}
    \centering
    \includegraphics[width=\textwidth]{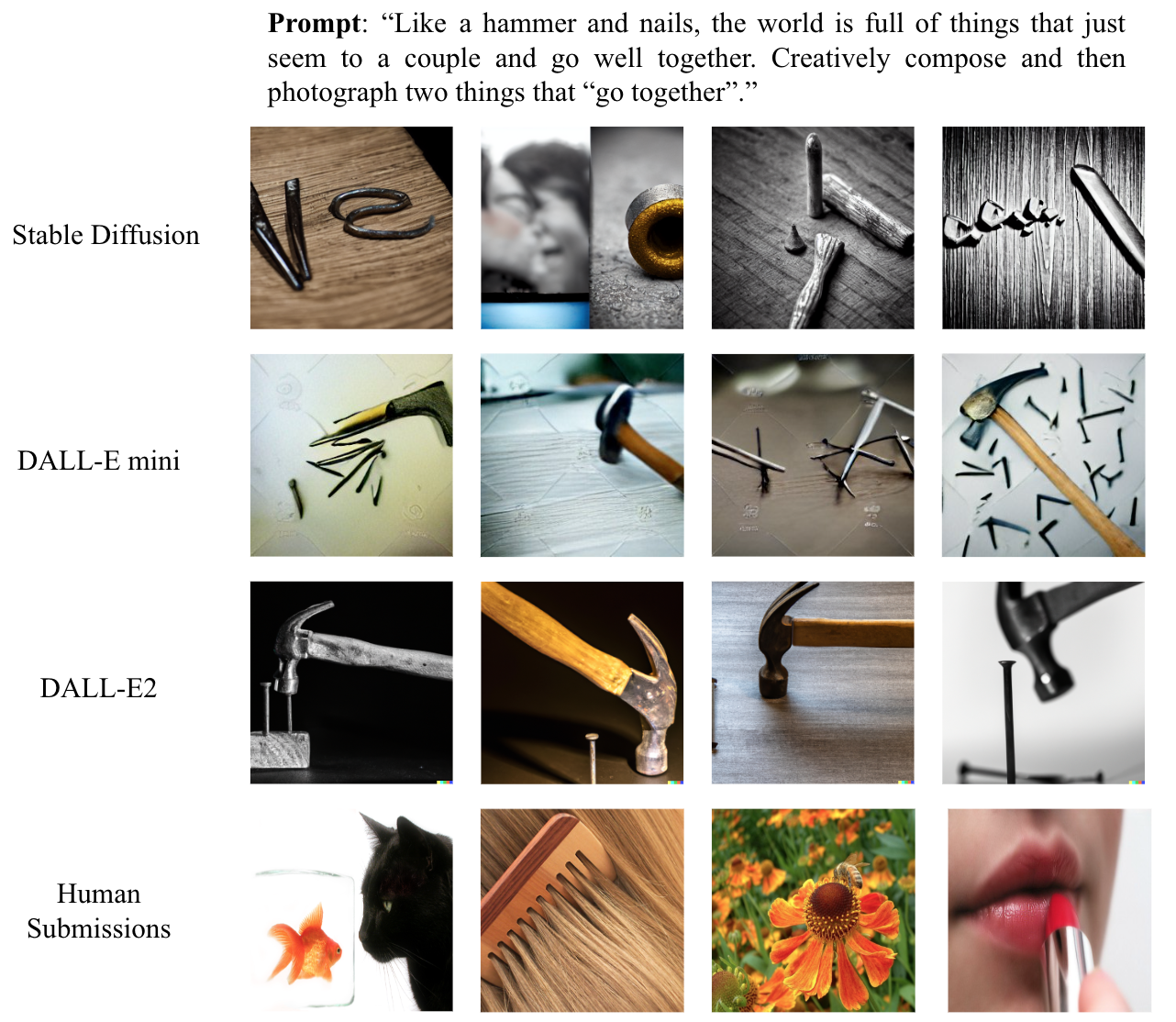}
    \caption{In this case the prompt suggests an example that participants can take figuratively to come up with new ideas, but instead diffusion models apparently understand only the literal meaning.}
    \label{fig:example_nails}
\end{figure}

Here we present some instances that we believe showcase one of the current strongest limitations of DGMs: their dependence on objective, fine-grained descriptions of the image in order to generate high-quality, semantically-similar images.
To illustrate this limitation, we feed the model with prompts taken from the instructions given by the organisers of photographic contests at \url{https://www.dpchallenge.com} to participants, and compare the outcome of some popular diffusion models against human submissions.
As it can be seen from Figure \ref{fig:example_beginning} and Figure \ref{fig:example_nails}, humans naturally understand figurative speech and creatively interpret the prompt in different ways.
Diffusion models seem to struggle to go beyond the literal meaning.

\section{Wasserstein distance between datasets}
\label{append: eom}

We chose the Wasserstein distance to provide an intuition about how much it would require to convert one distribution into another, and compute it for each pair of datasets in the three variables considered (sentimentality, subjectiveness and concreteness), as show in Table~\ref{tab:eom}.
The Wasserstein distance between two distributions $u$ and $v$ can be computed as:
\begin{equation}
    W_1(u, v) = \underset{\pi \in \Gamma (u, v)}{\mathrm{inf}} \int_{\mathbb{R} \times \mathbb{R}}|x - y|d\pi(x, y)
\end{equation}

where $\Gamma$ is the set of (probability) distributions on $\mathbb{R} \times \mathbb{R}$ whose marginals are $u$ and $v$ on the first and second factors respectively.

We find that the average distance to shift from any distribution to another within datasets related to art is smaller ($\mu(W_1)=2.71e^{-3}$, $\sigma^2(W_1)=0.6e^{-3}$) than that required to transform it into either MS-COCO, Conceptual Captions, Captions12M or Laion-Aesthetics.
An analogous effect is observed within these datasets ($\mu(W_1)=3.16e^{-3}$, $\sigma^2(W_1)=1.38e^{-3}$).

\begin{table}
    \centering
    \resizebox{\textwidth}{!}{%
    \begin{tabular}{c||c|cccccc}
         & & Artemis & SemArt & MS-COCO & Conc. Caps. & Captions12M & Laion-Aesth. \\
         \hline
         \hline
        \multirow{8}{*}{\rotatebox[origin=c]{90}{Sentiment}}
         & & & & & & \\
        & AVA & 3.53e-3 & 2.34e-3 & 9.51e-3 & 6.08e-3 & 4.24e-3 & 7.41e-3 \\
        & Artemis & - & 2.25e-3 & 1.23e-2 & 9.15e-3 & 7.12e-3 & 1.04e-2 \\
        & SemArt  &  & - & 1.17e-2 & 8.09e-2 & 6.14e-3 & 9.4e-3 \\
        & MS-COCO & & & -& 3.7e-3 & 5.76e-3 & 2.61e-3 \\
        & Conc. Caps. & & & & -& 2.23e-3 & 1.37e-3 \\
        & Captions12M & & & & & - & 3.3e-3 \\
        & \\
        \hline
         \hline
        \multirow{8}{*}{\rotatebox{90}{Subjectivity}} & & & & & & & \\
        & AVA  & 1.9e-3 & 4.55e-3 & 5.25e-3 & 5.49e-3 & 3.73e-3 & 6.98e-3 \\
        & Artemis & - & 4.1e-3 & 6.29e-3 & 6.76e-3 & 4.77e-3 & 8.24e-3 \\
        & SemArt  &  & - & 8.83e-3 & 8.78e-3 & 6.79e-3 & 9.78e-3 \\
        & MS-COCO & & & -& 1.17e-3 & 2.48e-3 & 2.43e-3 \\
        & Conc. Caps. & & & & -& 2.25e-3 & 1.55e-3 \\
        & Captions12M & & & & & - & 3.47e-3 \\
        & \\
        \hline
        \hline
        \multirow{8}{*}{\rotatebox{90}{Concreteness}} & & & & & & & \\
        & AVA  & 1.46e-3 & 1.46e-3 & 4.00e-3 & 3.69e-3 & 3.90e-3 & 5.8e-3 \\
        & Artemis & - & 1.09e-3 & 3.69e-3 & 3.13e-3 & 3.35e-3 & 5.55e-3 \\
        & SemArt &  & - & 2.76e-3 & 2.45e-3 & 2.62e-3  & 4.55e-3 \\
        & MS-COCO & & & -& 8.37e-4 & 1.02e-3 & 1.99e-3 \\
        & Conc. Caps. & & & & -& 4.95e-4 & 2.43e-3 \\
        & Captions12M & & & & & - & 2.27e-3 \\
        & \\
        \hline
        \hline
    \end{tabular}}
    \caption{Wasserstein distance ($W_1$) computed between each pair of datasets and language property considered in the study.}
    \label{tab:eom}
\end{table}

\end{document}